\documentclass[runningheads,a4paper]{llncs}
\usepackage{amssymb}
\usepackage{graphicx}
\usepackage{url}
\usepackage{algorithm}
\usepackage[noend]{algorithmic}
\usepackage{enumerate}
\usepackage{rotating}
\usepackage{multirow}
\usepackage{bigstrut}

\urldef{\mailsa}\path|{daniel.harabor, philip.kilby}@nicta.com.au|

\begin{document}

\mainmatter  

\title{Informed Heuristics For Guiding Stem-And-Cycle Ejection Chains}

\author{Daniel Harabor \and Philip Kilby}

\institute{NICTA and The Australian National University\\
7 London Circuit, Canberra, 2601, Australia\\
\mailsa 
}

\toctitle{Informed Heuristics For Guiding Stem-And-Cycle Ejection Chains}
\tocauthor{Daniel Harabor}
\tocauthor{Philip Kilby}
\maketitle

\begin{abstract}
  The state of the art in local search for the Traveling Salesman Problem is dominated by
  ejection chain methods utilising the Stem-and-Cycle reference structure.
  Though effective such algorithms employ very little information in their successor selection 
  strategy, typically seeking only to minimise the cost of a move.
  We propose an alternative approach inspired from the AI literature and show
  how an admissible heuristic can be used to guide successor selection.
  We undertake an empirical analysis and demonstrate that this technique often produces better results 
  than less informed strategies albeit at the cost of running in higher polynomial time.
\end{abstract}

\section{Introduction}
Simple to state yet difficult to solve, the Traveling Salesman Problem (TSP) is one of the oldest and most widely studied in 
the field of combinatorial optimisation.
The canonical (or symmetric) TSP requires finding a minimum cost tour of all nodes in a graph $G = (V, E)$ where $V = \lbrace v_{1},
\ldots, v_{n} \rbrace$ is the set of all nodes and $E = \lbrace (v_{i}, v_{j}) | v_{i}, v_{j} \in V, i \neq j \rbrace$
is a set of edges connecting them such that cost (or weight) of each edge $c_{ij} = c_{ji}$ and is non-negative for all
$(v_{i}, v_{j}) \in E$.

Despite receiving much attention the TSP has resisted attempts at developing a polynomial time algorithm to solve it
exactly; a situation which is unlikely to change unless $P = NP$ \cite{garey79}.
In response many researchers began focusing their attention on local search methods in order to produce approximate 
solutions in low polynomial time.
By far the most successful of these efforts is the seminal Lin-Kernighan algorithm \cite{lk73}, variations on which
have dominated the area for almost three decades.
Recently however it has been shown that alternative techniques, using the Subpath Ejection Chain (or SEC) procedure,
are able to compete with and in many cases outperform Lin-Kernighan \cite{rego98b,gutin02,gamboa06}.

SEC proceeds by identifying the \emph{Stem-and-Cycle reference structure}; a spanning subgraph of $G$ which is similar to but 
not quite a tour (yet from which a tour can always be immediately generated).
Using special transition rules one configuration of the reference structure is transformed into another such that a 
compound neighbourhood of reachable tours is created and from this neighbourhood a cost improving tour is selected.
The power of the technique can be explained, in part, by a connectivity result due to Glover \cite{glover96} 
which shows that the neighbourhood generated using Stem-and-Cycle is a super-set of that generated by Lin-Kernighan.

At the heart of the SEC procedure is a simple policy which extends the ejection 
chain by always choosing to expand the neighbouring configuration with minimum transition cost \cite{glover96}.
While effective it has been suggested that the nearest-neighbour strategy fails to exploit the geometric structure of 
the TSP and tends to perform poorly on problem instances where nodes appear in clusters \cite{gutin02}.
In the AI literature such difficulties are often overcome by extracting knowledge from the problem domain.
Known as \emph{informed heuristics} these methods estimate the cost of all remaining steps required to solve the problem. 
This is in stark contrast to the nearest-neighbour strategy which only considers one step at a time.

We develop an alternative move operator which uses an informed heuristic to rank neighbouring configurations 
according to their ``distance'' from an apriori unknown target tour. 
We compute a 1-tree lower-bound \cite{karp70} on the cost of the target tour and work on the thesis 
that this heuristic will guide our algorithm to neighbourhoods which contain lower 
cost tours and in relatively fewer steps than would otherwise be reachable using SEC. 

Our motivation is to show that on many smaller TSP instances ejection chains guided by informed heuristics are 
able to achieve better results than alternatives based on the nearest-neighbour strategy.
Smaller TSPs are of interest because they often appear in many other areas of combinatorial 
optimisation. For example, in the Vehicle Routing Problem the number of requests which may be assigned to any one
route is necessarily small (usually less than 100) due to physical limits on the capacity and range of each truck.

\section{Subpath Ejection Chains using Stem-and-Cycle}
\label{sec-stemandcycle}
\begin{figure}[htbp]
	\vspace{-4pt}
        \begin{center}
                        \includegraphics[scale=0.33, trim = 20mm 5mm 20mm 0mm]{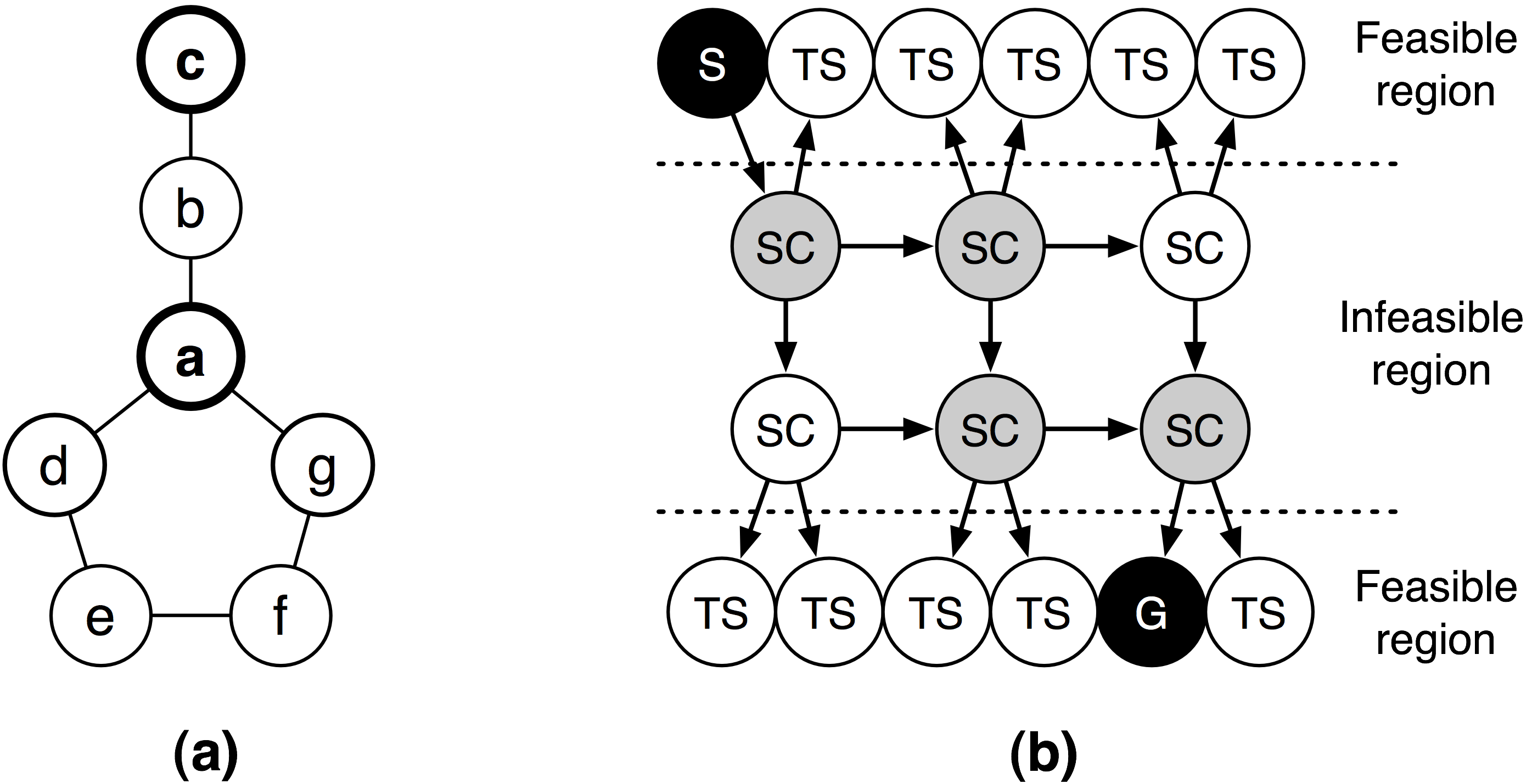}
        \end{center}
	\vspace{-7pt}
        \caption{\textbf{(a)} The Stem-and-Cycle reference structure. Node \emph{c} is the tip, node \emph{a} the root and
		nodes \emph{d} and \emph{g} the left and right subroots respectively. 
		\textbf{(b)}. SEC begins at tour S and creates a 4-level ejection chain (each level is denoted by a gray node) 
		before choosing G as the best reachable trial solution (TS).
	}
        \label{fig-scdiag}
	\vspace{-7pt}
\end{figure}

The Stem-And-Cycle reference structure, as shown in Figure \ref{fig-scdiag}(a), is a spanning subgraph of $G$ that
consists of a set of nodes $ST = \lbrace v_t, \ldots, v_{r} \rbrace$ forming a simple path called the \emph{stem}
which is connected to another set of nodes $CY = \lbrace v_{r}, v_{s_{1}}, \ldots,v_{s_{2}}, v_{r} \rbrace$ termed the \emph{cycle}.
The node connecting the stem to the cycle, $v_{r}$, is called the root while the node at the other end of the 
stem, $v_{t}$, is known as the tip.
$v_{r}$ must always be connected to at least two unique nodes on the cycle $v_{s_{1}}$ and $v_{s_{2}}$ which are
respectively designated the left and right subroots.

The main idea behind Stem-and-Cycle (whose theoretical underpinnings may be found in \cite{glover96})
is to search for a cost improving tour by re-arranging subpaths within the reference structure. 
This is achieved by defining a series of transition rules, or moves, that transform one Stem-and-Cycle configuration into
another.
Each move adds an edge $(v_{t}, v_{i})$ that connects the tip to a node on the stem or cycle
and deletes an adjacent edge $(v_{i}, v_{j})$ such that the resulting spanning graph remains connected. 
When $v_{t} = v_{r}$ the structure is said to be degenerate and forms a valid tour.
Further, although the Stem-and-Cycle structure is not itself a feasible solution to the TSP it is always possible to 
generate two such tours or \emph{trial solutions}, by adding an edge $(v_{t}, v_{s_{1}})$ or $(v_{t}, v_{s_{2}})$
and deleting the edge $(v_{s_{1}}, v_{r})$ or $(v_{s_{2}}, v_{r})$ as appropriate.

The Stem-and-Cycle reference structure is used as the basis for the Subpath Ejection Chain (or SEC) 
move operator. 
We briefly outline it here due to its central importance in developing other ideas in this paper; a 
full description, including pseudocode, is given in \cite{rego98b}.

In broad terms SEC explores a neighbourhood by producing a series of interrelated moves that generate one 
Stem-and-Cycle configuration from another in a chain-like fashion.
Each move constitutes a new level of the ejection chain and at each level it is necessary to evaluate the set of all
legitimate Stem-and-Cycle configurations in order to select one which will further extend the chain.
When the ejection chain has reached a maximal size the best trial solution reachable from any of its levels
is selected as the new start tour.
The process is repeated until SEC reaches a local minimum where no cost improving tour can be found.
Figure \ref{fig-scdiag}(b) serves to illustrate this idea. 

In order to guide SEC toward promising regions of the search space two important constraints are applied:

\begin{enumerate}
	\item{Each added edge is marked interchangeably as either black or white.
	  White edges are assumed to belong to a target tour into which we seek to transform
	  the starting tour. Thus, white edges, once added, cannot be subsequently deleted. 
	  Black edges are assumed to not belong to the target tour and may be deleted as part of a future move.
	  }
	\item{
	  Each deleted edge is marked ``tabu'' and cannot appear in a subsequent level of the ejection chain.
	  }
\end{enumerate}

The first constraint seeks to focus the search toward an apriori unknown target tour which may or may not be cost improving.
It also serves to limit the length of the ejection chain to a maximum of $2n$ levels, where $n = |V|$.
The second constraint prevents the search from cyclical backtracking.
Any move which violates one or both constraints is deemed to be illegitimate and not considered.

\section{Growing Ejection Chains Using Heuristic Search}
The SEC algorithm searches for cost improving tours by creating a series of multi-level ejection chains.
At each level of a chain the algorithm generates a set $\Gamma$ of up to $2n$ legitimate Stem-and-Cycle configurations
\cite{glover96}.
Each $x \in \Gamma$ is evaluated using $NN$, a nearest-neighbour heuristic function modeled as:

$$ NN(x) = ejectionValue + min(trialValue) $$ 

The first term, $ejectionValue$, refers to the cost of the move generating $x$ from the current Stem-and-Cycle
configuration. 
The second term, $min(trialValue$), is the minimum cost associated with generating one of two the trial solutions
reachable from $x$.
Since each $NN(x) | x \in \Gamma$ requires constant time to evaluate a maximally sized ejection chain
with up to $2n$ levels may be constructed in no more than $O(4n^2)$ time.

Although SEC runs very fast its nearest-neighbour heuristic is not very accurate. 
In particular, there is a risk that SEC will choose a successor which appears promising but ultimately
leads to a section of the search space that contains no cost improving tour.
To address this issue we propose an alternative move operator that is based, in part, on a well known 
algorithm from the heuristic search literature: A* \cite{hart68}.
Developed in the context of the single-source shortest-path problem, A* makes use of an informed heuristic function 
that evaluates nodes in the search space by computing a lower-bound on the remaining distance to reach a known goal node.
In the canonical implementation the final or \emph{f-cost} associated with an arbitrary node $x$ is given by the function:
$$f(x) = g(x) + h(x)$$
where $h$ is the heuristic cost and $g$ is the weight of the path from the start node to $x$. 
Provided $h$ is \emph{admissible} (i.e. it never overestimates the distance to the goal) this approach is guaranteed to 
find a shortest path to the goal if one exists. 
Applying such an algorithm to solve the TSP is not straightforward however. 
In particular, it is difficult to measure the ``distance'' between an arbitrary tour and the optimal tour as the latter is not known apriori.
Another difficulty arises from the fact that A* is a systematic tree-search algorithm; when applied to the TSP it exhibits exponential space complexity requirements
which make it infeasible for solving all but the most trivial instances \cite{pohl77,pearl84}.
In the remainder of this section we will address each of these problems in turn and in the process develop a new Stem-and-Cycle move operator
that constructs ejection chains with the help of a an informed heuristic function.

\subsection{Problem Formulation}

We begin by recalling that every ejection chain created by the Stem-and-Cycle move operator must respect a set of edge
exclusion constraints (namely those edges which have been previously deleted and cannot be re-added) and a set of edge inclusion
constraints (namely those edges marked as white which cannot be deleted).
Combined, these constraints guide the ejection chain toward a particular but apriori unknown tour.
We designate this tour $g$ and the associated start tour, from which the algorithm begins, $s$.
Let these tours correspond to A*'s goal node and start node respectively.

We define our search space as an undirected meta-graph $G_{m} = (V_{m}, E_{m})$ where $V_{m}$ represents the set of nodes
in the graph and $E_{m}$ the set of edges.
Let each node $x \in V_{m}$ represent a valid Stem-and-Cycle configuration.
We will say that an edge $(x, y) \in E_{m}$ exists between two nodes, $x$ and $y$, if the Stem-and-Cycle move operator
can induce $y$ from $x$ (or vice versa) in a single move. 
The weight of each edge is thus equal to the cost of the associated Stem-and-Cycle move that induces one of
its endpoints from the other. 
Under this formulation $s, g \in V_{m}$ and our goal is to find a path $\prod(s, g)$ connecting them
such that the weight of $g$ is minimised:

$$ \prod(s, g) : min(g) $$

The existence of the path is guaranteed by a result due to Glover \cite{glover96} which shows that Stem-and-Cycle can generate a tour from
any other using not more than $2n$ moves. 

\subsection{Modeling The Cost Function}
\label{sec-costfunction}
Recall that A* ranks nodes by computing a final value which is the sum of two cost
functions: \emph{g} and \emph{h}. 
We proceed by developing TSP-specific analogues for each of these.

First we consider the problem of calculating a \emph{g-cost} which requires exactly measuring the distance from $s$ to 
some candidate successor node $c \in V_{m}$.
To achieve this we can simply take the sum of the weights of all edges designated white appearing in $c$.
We reason that a node which contains more white edges is in some sense ``closer'' to $g$ than one containing fewer white edges.

Meanwhile, calculating a \emph{h-cost} is analogous to computing a lower-bound on the cost of all white edges 
which have yet to be added in order to transform $c$ into $g$.
We accomplish this by taking the weight of all non-white edges in a \emph{constrained 1-tree}.

\begin{figure}[htbp]
	\vspace{-4pt}
        \begin{center}
                        \includegraphics[scale=0.35, trim = 20mm 5mm 20mm 0mm]{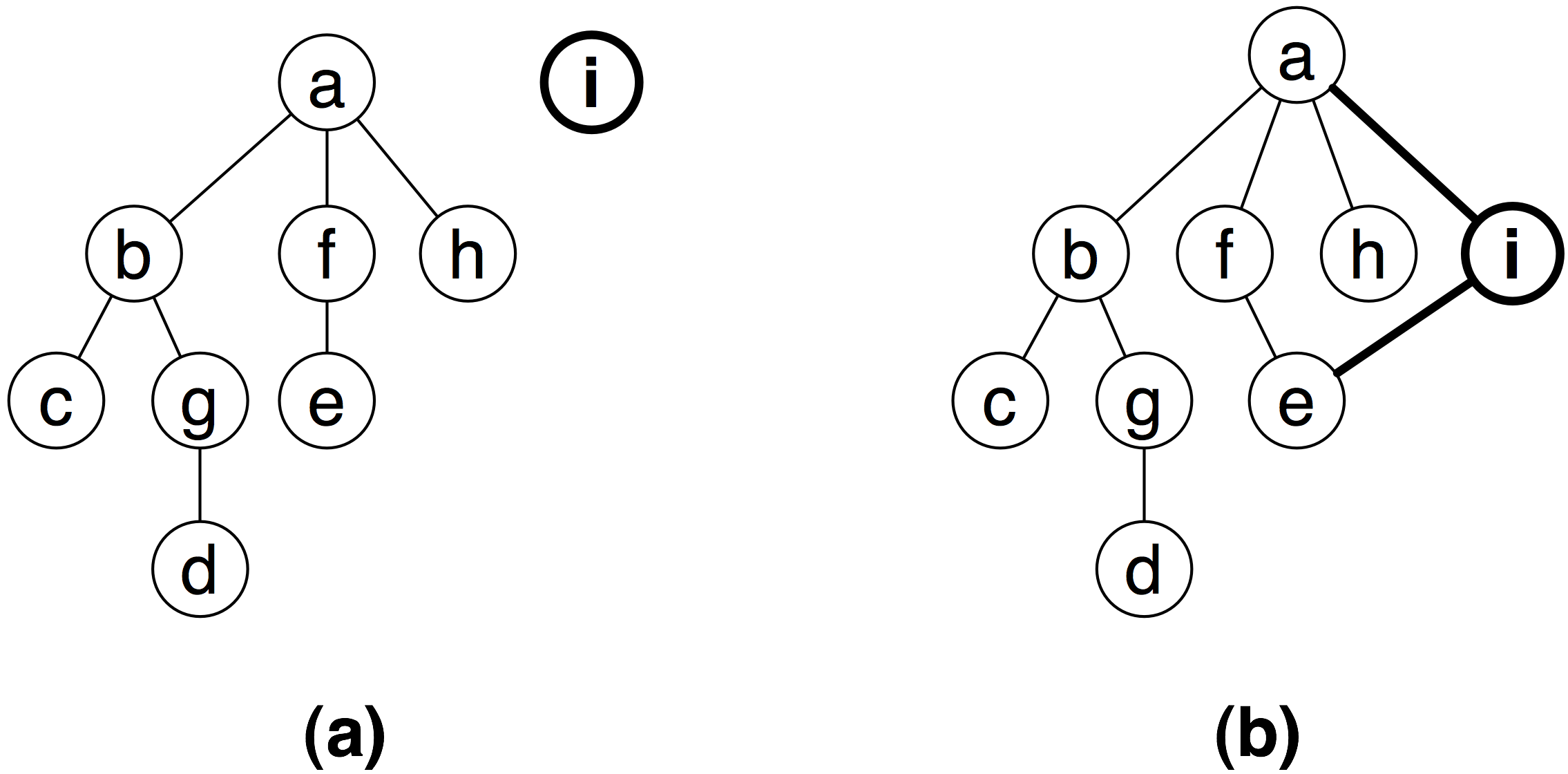}
        \end{center}
	\vspace{-7pt}
        \caption{ 
			\textbf{(a)} Begin the 1-tree by first constructing an MST of all tour nodes except one (here, node \emph{i}).
			\textbf{(b)} Complete the 1-tree by connecting node $i$ to the rest of the spanning tree via 
				the two smallest weight edges incident with it (here, edges $(i, a)$ and $(i, e)$).
	}
        \label{fig-1tree}
	\vspace{-7pt}
\end{figure}

Conceptually similar to the canonical 1-tree \cite{karp70} (the construction of which we outline in Figure \ref{fig-1tree})
the constrained 1-tree differs in that it contains all edges from a given include set and none of the edges from a given exclude set.
Algorithm \ref{alg-c1tree} describes the steps necessary to create this data structure.
Note that this idea was originally suggested by Pohl \cite{pohl77} but not described in any detail.
As such we believe this work is the first elaborate study of the constrained 1-tree. 

\begin{algorithm}
  \caption{Constrained 1-Tree}
  \label{alg-c1tree}
  \begin{algorithmic}[1]
	\newcommand{\algorithmicprocedure}{\textbf{procedure}\ }
	\newcommand{\algorithmicfunction}{\textbf{function}\ }
	\newcommand{\algorithmicprocend}{\textbf{end procedure}\ }
	\newcommand{\algorithmicfunctionend}{\textbf{end function}\ }
	
	\REQUIRE $G = (V, E)$, $ExcludedEdges$, $IncludedEdges$
	\STATE $v_{1} \leftarrow $getNodeAtIndex($V$,$0$)
	\STATE $mstroot \leftarrow$ getNodeAtIndex($V$, $1$)
	\STATE $Neighbours$ $\leftarrow$ $V - \lbrace v_{1} \cup mstroot \rbrace$
	\STATE $Heap \leftarrow$ initHeap($Neighbours$)
	\STATE processNeighbours($mstroot$)
	\WHILE{$Neighbours \neq \emptyset$}
		\STATE $n \leftarrow$ min($Heap$)
		\IF{parent($n$) $= \varnothing$ \textbf{or} $n = \varnothing$}
			\RETURN $\varnothing$
		\ENDIF
		\IF{makeEdge(parent($p$), $n$) $\in ExcludedEdges$}
			\RETURN $\varnothing$
		\ENDIF
		\STATE parent($n$).addChild($n$)
		\STATE $Neighbours \leftarrow \lbrace Neighbours - n \rbrace$
		\STATE processNeighbours($n$)
	\ENDWHILE
	\RETURN make1Tree($mstroot$, $v_{1}$)
	\newline \newline
	\algorithmicprocedure processNeighbours($n$)
	\FORALL{$x \in Neighbours$}
	  \STATE e $\leftarrow$ makeEdge($n$, $x$)
	  \IF{$e \in IncludedEdges$}
	  	\STATE Heap.priority($x$, 0)
		\STATE parent($x$) $\leftarrow n$
	  \ELSE
		\IF{$e \not \in ExcludedEdges$ \textbf{and} priority($x$) $<$ cost($e$)}
		  \STATE Heap.priority($x$, cost($e$))
		  \STATE parent($x$) $\leftarrow n$
		\ENDIF	
	  \ENDIF
	\ENDFOR
  \end{algorithmic}
\end{algorithm}

Our implementation takes a set of nodes $V = \lbrace v_{1}, \ldots, v_{n} \rbrace$ and proceeds by creating
a MST on the subset of nodes $\lbrace v_{2}, \ldots, v_{n} \rbrace$.
However we augment the basic MST construction algorithm of Prim (as described in \cite{cormen01}) such that it respects 
all edge constraints to which the move operator is subject.
We make two changes:

\begin{enumerate}
	\item{In lines 19-21 of Algorithm \ref{alg-c1tree}: white edges are added to the heap are given the highest
	  priority.
	  This ensures that all such edges are immediately selected for inclusion into the MST. }
  \item{In lines 23-25 of Algorithm \ref{alg-c1tree}: before any non-white edge is added to the heap we make sure it is not on the excluded edges list.
	  If it is, we immediately discard it from further consideration.
	  }
\end{enumerate}

We also apply the following rules at line 16 of Algorithm \ref{alg-c1tree} (where we connect the excluded node $v_{1}$ to the
MST and complete the 1-tree):

  \begin{enumerate}
	\item{Any white edge incident with node $v_{1}$ which appears in the included edges list is automatically selected, regardless of cost. }
	\item{Any non-white edge incident with node $v_{1}$ which appears in the excluded edges list is automatically discarded, regardless of cost.}
	\item{The remaining edges incident with node $v_{1}$ are ranked by their weight. 
	  From this set we always select the one with smallest weight.}
  \end{enumerate}

\subsection{On The Admissibility Of The Constrained 1-Tree Heuristic}
The canonical 1-tree always lower-bounds the optimal tour and thus the the goal configuration $g$.
However, it is not immediate that the constrained 1-tree, as constructed by Algorithm 1, is also a lower-bound on $g$.
In particular we would like to show that the heuristic function $h$, which is based on the sum cost of non-white
edges in a constrained 1-tree, is a lower-bound on the sum cost of any as-yet-unknown set of white edges which 
appear in the goal tour $g$.
This property is required to ensure the admissibility of the heuristic function $h$.

\begin{lemma}
  \label{lemma-c1texists}
  For all $c \in V_{m}$ it is always possible to construct a constrained 1-tree.
\end{lemma}
\begin{proof}
To prove this claim it is sufficient to observe that every Stem-and-Cycle configuration, including the optimal
tour, is itself a 1-tree. 
\qed
\end{proof}

\begin{lemma}
  \label{lemma-mstweight}
  The constrained minimum spanning tree associated with an arbitrary node $c \in V_{m}$ is a lower-bound on the goal
  configuration $g$.
\end{lemma}
\begin{proof}
  We proceed with a structural proof by induction on the spanning tree constructed for $c$.
  First we will analyse two distinct possibilities for the base case: $c = s$ or $c = g$.
  We then proceed with an analysis of the general case in which $c$ is subject to an arbitrary number of 
  edge inclusion and edge exclusion constraints. 
  In each case we will show that Algorithm \ref{alg-c1tree} produces a spanning tree which is a lower-bound on $g$.
\begin{description}
  \item[Base Case 1:]{$c = s$.
	In this case there are no edge inclusion or exclusion constraints.
	Algorithm \ref{alg-c1tree} mirrors the behaviour of Prim's algorithm and thus creates an unconstrained 
	spanning tree of minimum weight.
	Clearly such a structure is a lower-bound on $g$.
	}
  \item[Base Case 2:]{$c = g$. In this case every edge in $c$ is white and appears in the set of edges 
	that must be included.
	Algorithm \ref{alg-c1tree} constructs the constrained MST by looking for minimum cost edges that connect a leaf of 
	the spanning tree to some other node that is not yet in the tree.
	If any edge from this candidate set is white it is assigned the highest priority which guarantees it 
	will be selected the next time the spanning tree is extended. 
	Since every node in $c$ is incident with at least one white edge it follows that each node is added to the 
	spanning tree by way of a white edge.
	As the constrained MST contains only white edges its weight must be a lower-bound on $g$.
	}
  \item[Inductive Case:]{
	$c$ contains some number of white and non-white edges.
	We know from Base Case 2 that every white edge appearing in $g$	also appears in the constrained MST 
	associated with $c$.
	It remains to show that the set of non-white edges appearing in the constrained MST is minimum. 

	When considering a non-white edge Algorithm \ref{alg-c1tree} will rank candidates that have not been explicitly
	excluded by their weight and
	always select for inclusion into the spanning tree the edge with smallest weight.
	Consequently, no edges which have been deleted can appear in the spanning tree and of the remaining candidate edges
	only those edges of least cost which do not induce a cycle are selected when extending the tree.
	The set of non-white edges included in the spanning tree is thus minimum and must be a lower-bound on the weight of any 
	white edges which later appear in $g$.
	Thus the weight of the constrained MST is a lower-bound on $g$.
	}
\end{description}
\qed
\end{proof}

\begin{theorem}
  \label{thm-c1tadmissible}
  The weight of a constrained 1-tree associated with some arbitrary node $c \in V_{m}$ is a lower-bound on the
weight of the goal configuration $g$. 
\end{theorem}
\begin{proof}
  We recall that a constrained 1-tree on the set of nodes $\lbrace v_{1},\ldots,v_{n} \rbrace \in c$ is a two-step process 
  requiring the union of a constrained MST on the nodes $\lbrace v_{2},\ldots,v_{n} \rbrace$ with the two least-cost 
  (possibly white) valid edges incident with $v_{1}$. 
  We know from Lemma \ref{lemma-mstweight} that the weight of the constrained MST is a lower-bound on $g$. 
  It remains to show that the two edges connecting $v_{1}$ to the spanning tree are a lower-bound on the cost of two
  edges incident with $v_{1}$ in the goal configuration $g$.

  This operation follows the edge selection policy outlined in Section \ref{sec-costfunction}; 
  i.e. we preference white edges above all others, regardless of cost, disconsider
  excluded edges and rank the remaining edges by weight.
  Thus the two edges selected to connect $v_{1}$ to the constrained MST must be a lower-bound on the cost of the two edges 
  incident with $v_{1}$ in $g$.
  From this we conclude that the constrained 1-tree constructed by Algorithm \ref{alg-c1tree} is itself
   a lower-bound on $g$.
\qed
\end{proof}

Theorem \ref{thm-c1tadmissible} is sufficient to show that a $h$ is admissible.
Furthermore, each generated node evaluated by $h$ will be subject to a larger set of edge constraints than its parent.
From this it is readily shown that $h$ is also \emph{monotonic} and thus guaranteed to be non-increasing as the search approaches 
the target tour $g$.

\subsection{The Informed Subpath Ejection Chain Algorithm}
We have combined the Stem-and-Cycle reference structure with the constrained 1-tree lower-bound in order to develop a 
TSP-specific analogue of A*.
If we apply the canonical form of the algorithm to $G_{m}$, always expanding the node with the least lower-bound, 
we are guaranteed to find the optimal tour.
However, this operation is likely to be computationally intractable unless our heuristic function $h$ 
is able to compute a near-perfect estimate of the remaining distance to reach the optimal tour.
If this is not the case A* will expand an exponential number of nodes \cite{pearl84}.



Since we are interested in an approach that is guaranteed to run in low polynomial time we modify the
canonical A* algorithm by reducing the branching factor of each node to 1.
This makes our new algorithm analogous to SEC \cite{glover96,rego98b} however instead of always choosing
the neighbour with minimum transition cost as the next level of the ejection chain we instead
choose the neighbour which minimises the cost of our evaluation function $f$.
We term the resulting technique Informed SEC (or ISEC for short) and outline it in Algorithm \ref{alg-isec}.

\begin{algorithm}
  \caption{Informed Subpath Ejection Chain Algorithm (ISEC)}
  \label{alg-isec}
  \begin{algorithmic}[1]
	\REQUIRE $s$
	\STATE $next \leftarrow s$
	\STATE $bestTour \leftarrow$ minTrialSolution($s$)
	\WHILE{$next \neq \varnothing$}
	  \IF{minTrialSolution($next$) $< bestTour$}
	  	\STATE $bestTour \leftarrow$ minTrialSolution($next$);
	  \ENDIF
	  \STATE $\Gamma_{next} \leftarrow$ expand($next$);
	  \STATE $next \leftarrow \varnothing$
	  \FORALL{$x \in \Gamma_{next}$}
		\IF{$next = \varnothing$}
			\STATE $next \leftarrow x$
		\ELSE
		  \IF{f($x$) $<$ f($next$)}
		  	\STATE $next \leftarrow x$
		  \ENDIF
		\ENDIF
	  \ENDFOR
	\ENDWHILE
	\RETURN $bestTour$ 
  \end{algorithmic}
\end{algorithm}

\section{Complexity}
\label{sec-complexity}
We analyse the running time of the ISEC algorithm in terms of node expansions.
When a node $x \in G_{m}$ is expanded we generate the set $\Gamma_{x}$ which contains
all legitimate Stem-and-Cycle configurations reachable in a single move.
We then select $min(\Gamma_{x})$ as the next node to be expanded and the process
continues until a maximal size ejection chain has been built.
Since every second move adds to $x$ a white edge which cannot be deleted the maximum
number of nodes that ISEC can expand per ejection chain is $2n$.
Additionally, the size of $\Gamma_{x}$ is itself not more than $2n$ which gives us an $O(4n^2)$
bound on the total number of nodes that must be generated per ejection chain. 

For each node $y \in \Gamma_{x}$ we call Algorithm \ref{alg-c1tree} to create a constrained 1-tree.
Using a derivative of Prim's MST algorithm we first compute a constrained MST on the 
nodes $\lbrace v_{2}, \ldots, v_{n} \rbrace \in V$.
This is an $O((n-1)^2 \times log_{2}(n-1))$ operation on a complete graph such as the Symmetric TSP.
To finish the 1-tree we connect node $v_{1} \in V$ to the constrained MST via its two least-cost edges.
However, we have to exclude from consideration any edges which have been previously deleted
and prioritise any edges which have been marked white.
Finding two edges which satisfy these constraints is an $O(n-1)$ operation, 
even if we pre-sort the the set of edges incident with $v_{1}$. 
Thus the total time required to compute a constrained 1-tree is $O((n-1)^2 \times \log_{2}(n-1) + (n-1))$

If we assume that each ejection chain constructed by ISEC will require a maximal number of node expansions
then the total running time of the algorithm is $O(4n^2\times [(n-1)^2 \times \log_2(n-1) + (n-1)])$.
We may roughly approximate this as $O(4n^4 \times log_{2}(n) + 4n^3)$.

Clearly the performance of ISEC is limited by the time required to compute a constrained 1-tree.
However, it is not strictly necessary to compute a new 1-tree for each generated node.
Consider the following argument:

Let $p, c \in V_{m}$ be two Stem-and-Cycle configurations such that $c$ is generated as a result of expanding $p$.
Such a move adds edge $e_{a}$ and deletes the edge $e_{d}$.
Next, let $\tau_{p}$ and $\tau_{c}$ be the associated constrained 1-trees which are constructed to evaluate $p$ and $c$ respectively.
We proceed with a discussion on the effect that the Stem-and-Cycle move generating $c$ from $p$ has on the structure of
$\tau_{c}$ with respect to $\tau_{p}$. 
In particular, we identify four distinct types of changes which may occur:

\begin{enumerate}[{Change} 1:]
  \item{
	$e_{a} \not \in \tau_{p}$ and $e_{d} \not \in \tau_{p}$. 
	In this case two possibilities exist, depending on whether $e_{a}$ is white or not.
	\begin{enumerate}[i{)}]
	  \item{If we suppose $e_{a}$ is not white then we are not forced to include in $\tau_{c}$ any edges which were 
		not already in $\tau_{p}$.
		Further, since $e_{d} \not \in \tau_{p}$ we do not have prevent any edges appearing in 
		$\tau_{c}$ which were present in $\tau_{p}$.
		Thus, $\tau_{p} = \tau_{c}$ and we do not have to generate a separate 1-tree for $c$. 
	  }
	  \item {
		If we suppose $e_{a}$ is white then we are forced to include it in $\tau_{c}$. 
		However as $e_{a} \not \in \tau_{p}$ we cannot re-use $\tau_{p}$ in place of $\tau_{c}$ and still retain the admissibility
		property of the heuristic function $h$.
		Thus we must generate $\tau_{c}$ from scratch.
	  }
	\end{enumerate}
	}

\item{
  \label{item-change2}
  $e_{a} \not \in \tau_{p}$ and $e_{d} \in \tau_{p}$. 
  In this case the deletion of $e_{d}$ divides $\tau_{p}$ into two separate subtrees which must be re-connected 
  with a least-cost edge that has not been previously deleted. 
  Finding such an edge is an operation requiring $O(n^2)$ time and thus asymptotically equivalent with computing
  $t_{c}$ from scratch. 
  The status of $e_{a}$ (black or white) is of no consequence.
}

\item{
  \label{item-change3}
  $e_{a} \in \tau_{p}$ and $e_{d} \not \in \tau_{p}$
  In this case two possibilities exist, depending on whether $e_{a}$ is white or not.
  \begin{enumerate}[i{)}]
	\item{
  \label{item-change3i}
	  If $e_{a}$ is a white edge its inclusion into $\tau_{c}$ is forced.
	  Consequently, Algorithm \ref{alg-c1tree} will produce a 1-tree such that $cost(\tau_{c}) \geq cost(\tau_{p})$.
	  However, as $e_{a} \in \tau_{p}$ it must be that $\tau_{p}$ respects the same set of constraints to which $\tau_{c}$ is
	  subject. 
	  Thus, $\tau_{p}$ is also a valid lower-bound for $c$ and we may use it in favour of generating $\tau_{c}$.
	}
	\item{
	\label{item-change3ii}
	  If $e_{a}$ is a black edge the same argument as before holds;
	  $\tau_{p}$ is a direct lower-bound for $c$ and we may use it in favour
	  of generating $\tau_{c}$.
	  }
  \end{enumerate}
  }

\item{
$e_{a} \in \tau_{p}$ and $e_{d} \in \tau_{p}$.
The same argument which we used for Change \ref{item-change2} is also applicable here.
The deletion of $e_{d}$ divides $\tau_{p}$ into two separate subtrees which must be re-connected with a least cost-edge.
This operation is asymptotically equivalent to generating $\tau_{c}$ from scratch.
The status of $e_{a}$ (black or white) is of no consequence.
}
\end{enumerate}

These results are interesting because they show that in half of all node generation scenarios it is not necessary
to compute a 1-tree.
In spite of this the theoretical running time of ISEC remains unchanged and in the worst case we still need to compute 
a constrained 1-tree for every node generated.
We have observed that in practice this scenario is unlikely; as our experimental results will show a speedup
of 50\% is not uncommon.
There are however some theoretical side-effects which merit discussion.
In particular, notice that in Change \ref{item-change3i} although $\tau_{c}$ can be shown to obey the
same set of constraints as $\tau_{p}$ its cost may infact be higher.
In such cases the accuracy of the heuristic function $h$ is diminished as not all constraints were taken
into consideration when computing $h(c)$.
We explore this issue empirically and show that in practice it is not a significant drawback.

\section{Experimental Setup}
We implemented and tested SEC (as described in \cite{rego98b}), ISEC and fast-ISEC (a variant of ISEC using the speedup
technique described in Section \ref{sec-complexity}) on 30 small-to-medium problems from TSPLIB
\cite{tsplib}. 
These ranged in size from 29 to 200 cities.
We generate 10 experiments per problem by randomly perturbing the initial order of cities.
We evaluate the performance of both algorithms on each of these 10 instances making for a total of 600 distinct experiments.
All tests were undertaken on an Intel Core2Duo machine with 2GB of RAM running OSX 10.5.7.

Our main objective is to compare the relative quality of successor nodes chosen during a neighbourhood search.
To this end we ran a single iteration of both SEC and ISEC and measured the length of the best corresponding
trial solution. 
Note that in the case of ISEC an iteration refers to the construction of a single ejection chain of maximal length.
In the case of SEC an iteration may consist of many such such ejection chains; 
the algorithm only terminates when a local minima is reached.

\section{Results}
Table \ref{table-quality} summarises the quality of computed solutions for the SEC, ISEC and fast-ISEC
algorithms. 
We evaluate algorithmic performance by taking the length of each solution found and 
measuring the relative percentage deviation from the optimal solution. 
In each case we give figures for the best and average solutions found (columns Min and Mean 
respectively). 
We also provide average running times 
for each algorithm (in columns CPU) which are taken as wall-clock time and measured in seconds.
Finally we evaluate the average quality of the starting tour (column STQ)
as measured in multiples of the optimal solution length.

\begin{table}
  \caption{Quality of solutions (\% above optimal) and CPU time (in seconds) for SEC, ISEC and fast-ISEC.}
  \label{table-quality}
  \centering
  \begin{tabular}{lcccccp{0.5em}cccp{0.5em}ccc}
   \hline
   \multirow{2}{*}{Problem} & \multirow{2}{*}{Optimal} &  \multirow{2}{*}{STQ} & \multicolumn{3}{c}{ISEC} & & \multicolumn{3}{c}{fISEC} & & \multicolumn{3}{c}{SEC} \\
  \cline{4-6}
  \cline{8-10}
  \cline{12-14}
   &  & &  Min & Mean & CPU &   & Min  & Mean & CPU &   & Min & Mean & CPU \\ 
  \hline
bayg29  & 1610 & 2.96 & 14.60 & 23.34 & 0.13 &    & 14.60 & 24.87 & 0.09 &    & 5.59 & 15.11 & 0.00 \\ 
  berlin52  & 7542 & 400.20 & 19.49 & 30.83 & 0.98 &    & 18.86 & 37.23 & 0.58 &    & 17.32 & 27.61 & 0.01 \\ 
  bier127  & 118282 & 521.52 & 28.58 & 35.14 & 23.97 &    & 27.50 & 35.92 & 12.88 &    & 27.68 & 35.88 & 0.09 \\ 
  ch130  & 6110 & 764.88 & 27.72 & 39.19 & 26.91 &    & 27.79 & 40.55 & 14.53 &    & 42.51 & 58.72 & 0.08 \\ 
  ch150  & 6528 & 811.33 & 21.71 & 43.09 & 46.57 &    & 32.10 & 50.97 & 24.51 &    & 62.40 & 82.20 & 0.11 \\ 
  d198  & 15780 & 1202.76 & 29.38 & 39.43 & 135.78 &    & 29.38 & 39.35 & 70.88 &    & 60.32 & 67.72 & 0.17 \\ 
  eil101  & 629 & 543.48 & 17.20 & 31.63 & 10.30 &    & 26.24 & 35.56 & 5.71 &    & 25.90 & 35.10 & 0.05 \\ 
  eil51  & 426 & 378.29 & 11.69 & 24.36 & 0.91 &    & 11.25 & 29.46 & 0.54 &    & 11.03 & 22.09 & 0.01 \\ 
  eil76  & 538 & 459.30 & 13.61 & 32.18 & 3.74 &    & 20.28 & 36.99 & 2.07 &    & 26.47 & 41.20 & 0.02 \\ 
  kroA100  & 21282 & 786.14 & 25.01 & 43.47 & 10.19 &    & 38.45 & 49.29 & 5.54 &    & 32.58 & 50.81 & 0.04 \\ 
  kroA150  & 26524 & 957.21 & 29.64 & 41.96 & 47.45 &    & 34.42 & 46.59 & 25.44 &    & 45.29 & 56.03 & 0.14 \\ 
  kroA200  & 29368 & 1165.43 & 38.66 & 47.27 & 150.71 &    & 38.83 & 50.28 & 77.80 &    & 53.40 & 81.68 & 0.18 \\ 
  kroB100  & 22141 & 754.03 & 22.62 & 35.92 & 10.26 &    & 22.71 & 39.09 & 5.63 &    & 37.71 & 57.51 & 0.05 \\ 
  kroB150  & 26130 & 1002.18 & 31.70 & 44.77 & 47.25 &    & 31.70 & 49.86 & 24.71 &    & 58.01 & 66.74 & 0.12 \\ 
  kroB200  & 29437 & 1140.09 & 45.67 & 54.03 & 145.55 &    & 38.80 & 49.75 & 77.06 &    & 64.23 & 83.93 & 0.20 \\ 
  kroC100  & 20749 & 823.89 & 31.58 & 42.36 & 10.36 &    & 18.79 & 44.53 & 5.68 &    & 38.71 & 55.75 & 0.04 \\ 
  kroD100  & 21294 & 776.11 & 31.97 & 39.82 & 10.49 &    & 33.29 & 48.22 & 5.59 &    & 37.33 & 51.13 & 0.04 \\ 
  kroE100  & 22068 & 756.53 & 29.91 & 46.35 & 10.44 &    & 27.99 & 46.89 & 5.59 &    & 45.20 & 73.82 & 0.04 \\ 
  lin105  & 14379 & 845.14 & 42.62 & 58.56 & 11.85 &    & 37.15 & 63.15 & 6.57 &    & 51.57 & 87.18 & 0.04 \\ 
  pr107  & 44303 & 1297.00 & 33.95 & 64.24 & 12.08 &    & 34.10 & 58.51 & 6.59 &    & 38.20 & 49.27 & 0.05 \\ 
  pr124  & 59030 & 1176.19 & 37.37 & 61.71 & 21.82 &    & 45.05 & 68.07 & 11.70 &    & 64.38 & 103.39 & 0.06 \\ 
  pr136  & 96772 & 832.38 & 27.40 & 42.08 & 30.18 &    & 37.80 & 47.77 & 16.42 &    & 49.36 & 68.92 & 0.07 \\ 
  pr144  & 58537 & 1361.50 & 40.13 & 74.17 & 37.10 &    & 39.21 & 66.39 & 20.24 &    & 65.06 & 90.74 & 0.09 \\ 
  pr152  & 73682 & 1398.02 & 49.02 & 63.52 & 45.64 &    & 35.76 & 59.54 & 25.10 &    & 48.74 & 70.63 & 0.13 \\ 
  pr76  & 108159 & 522.91 & 24.65 & 34.35 & 3.45 &    & 28.65 & 41.75 & 2.02 &    & 36.74 & 47.90 & 0.03 \\ 
  rat195  & 2323 & 968.21 & 33.77 & 41.01 & 126.33 &    & 35.69 & 42.24 & 66.77 &    & 80.39 & 95.76 & 0.19 \\ 
  rat99  & 1211 & 682.64 & 24.34 & 40.77 & 9.32 &    & 24.34 & 43.57 & 5.14 &    & 39.38 & 74.09 & 0.04 \\ 
  rd100  & 7910 & 709.22 & 24.36 & 39.70 & 10.45 &    & 26.62 & 39.76 & 5.65 &    & 38.23 & 57.64 & 0.04 \\ 
  st70  & 675 & 560.14 & 22.69 & 34.73 & 2.83 &    & 26.36 & 40.15 & 1.61 &    & 33.60 & 54.91 & 0.02 \\ 
  u159  & 42080 & 1059.40 & 46.24 & 54.21 & 53.80 &    & 39.20 & 54.18 & 28.59 &    & 72.48 & 93.99 & 0.12 \\ 
   \hline
   \multicolumn{2}{c}{Mean} & 821.97 & 29.24 & 43.47 & 35.23 & & 30.10 & 46.02 & 18.71 & & 43.66 & 61.91 & 0.08 \\
   \hline
   \vspace{-3em}
\end{tabular}
\end{table}

\begin{figure}[htbp]
        \begin{center}
                       \includegraphics[scale=0.40, trim = 20mm 17mm 20mm 5mm]{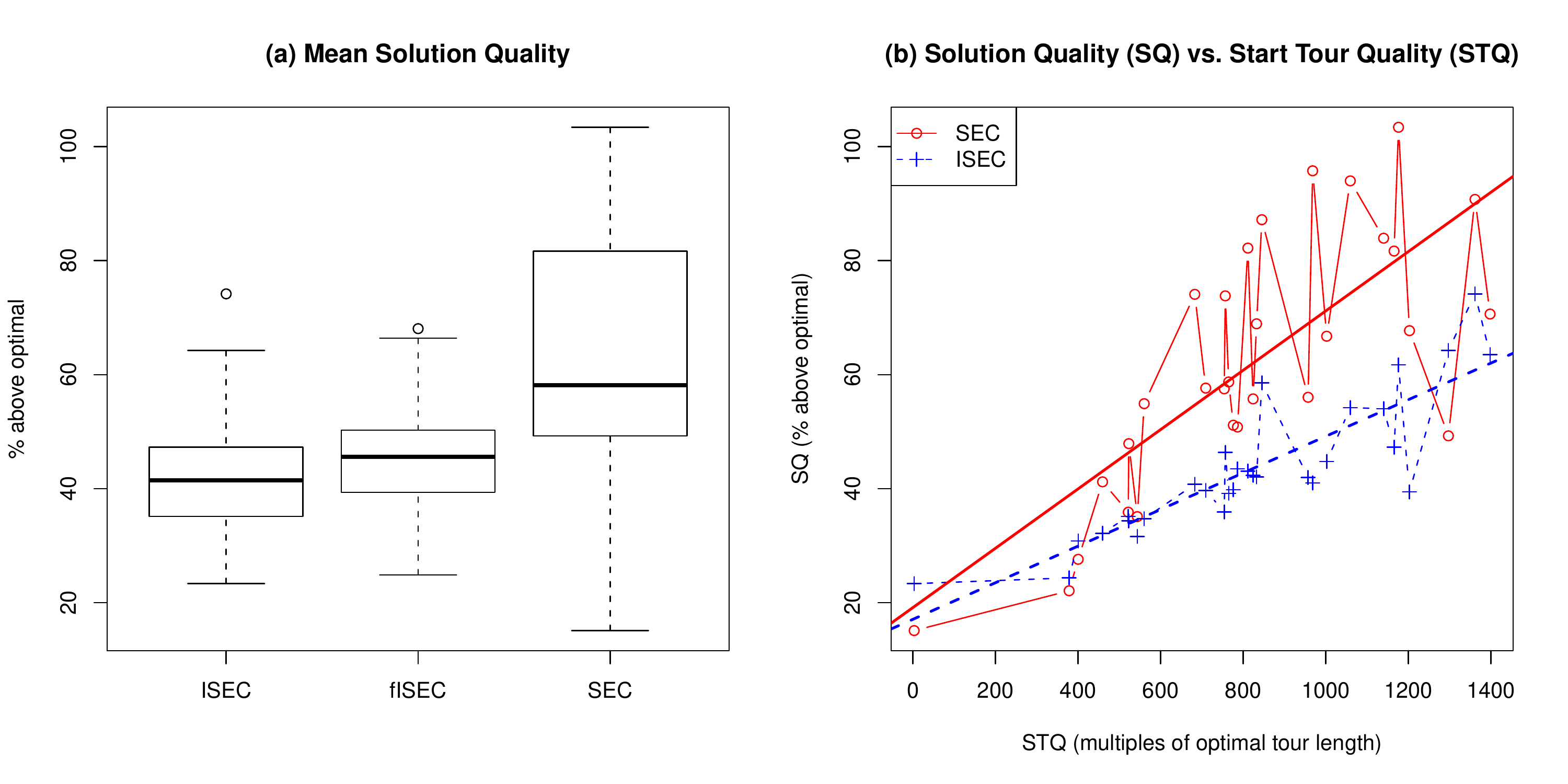}
        \end{center}
		\caption{\textbf{(a)} Variability of mean solution quality (measured as \% above optimal).
		\textbf{(b)} The effect of start tour quality on solution quality.}
        \label{fig-quality}
\end{figure}

Looking at the Table \ref{table-quality} we see that with respect to the quality of the 
average solution found ISEC outperforms SEC in 26 of the 30 problem instances.
In particular, the tours found by ISEC are on average 18.44\% closer to optimal.
A similar observation holds if we look at the quality of minimum solutions found.
As before, ISEC's performance dominates SEC -- this time in 25 of the 30 problem instances --
by an average of 14.4\%.
Similar results are seen in the case of fISEC however the average and minimum solution quality
improvements over SEC are slightly lower at 15.89\% and 13.56\% respectively.
As this is quite a coarse analysis we turn our attention to Figure \ref{fig-quality}(a) which shows 
the spread of results for mean solution quality.

Here we notice that the inter-quartile range of tours found by ISEC is just 11\% and includes
no solutions more than 46\% from optimal. 
Meanwhile SEC exhibits quite different behaviour; 
the inter-quartile range encompasses a very wide 39\% range and includes
tours which are up to 80\% above optimal. 
To explore this effect we turn our attention to Figure \ref{fig-quality}(b) in which we measure 
the effect of start tour quality on solution quality.

Using a simple least squares regression analysis we are able to see that the performance of both 
algorithms appears to be strongly correlated to the quality of the starting tour.
In particular, notice that as the ``distance'' between the optimal tour and the start tour
increases there is a corresponding drop in the mean quality of the best solution found.
This suggests that such problems are ``harder''  than others, requiring the ejection
of a greater number of edges before a good quality solution is found.
As before, ISEC performs much better than SEC almost across the board. 
This is directly attributable to our informed heuristic function which, it appears, is able to find
promising regions of the search space much more consistently than SEC's nearest-neighbour method.

\begin{figure}[htbp]
        \vspace{-2pt}
        \begin{center}
                       \includegraphics[scale=0.40, trim = 20mm 17mm 20mm 5mm]{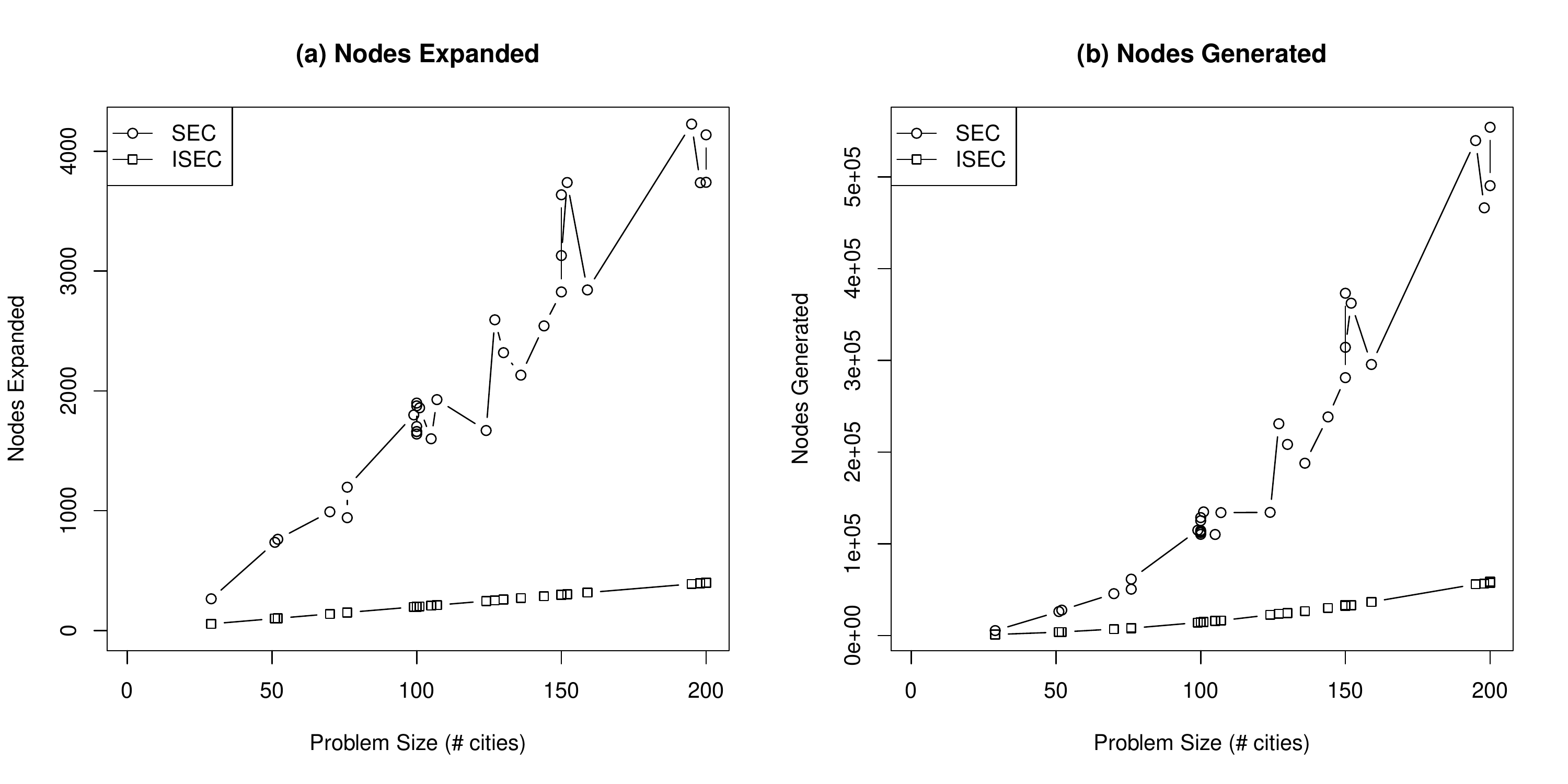}
        \end{center}
		\caption{Total search effort (SEC vs. ISEC).} 
        \label{fig-searcheffort}
        \vspace{-12pt}
\end{figure}

Next, we analyse the search effort of SEC and ISEC which we measure in terms of nodes expanded and
nodes generated.
Looking at Figure \ref{fig-searcheffort} we notice that in the case of both metrics the performance of
the two algorithms scales linearly with problem size. 
In the case of ISEC this is because the algorithm only constructs a single ejection chain; 
thus no more than $2n$ nodes can ever be expanded.
By comparison, SEC tends to expand (and consequently generate) an order of magnitude more nodes. 
From this we may infer that on average approximately 10 ejections chains must be constructed before the algorithm
reaches a local minima and terminates.

An interesting dichotomy appears when we contrast Figure \ref{fig-searcheffort} with the corresponding CPU times 
from Table \ref{table-quality}.
We see that although ISEC generates much fewer nodes than SEC it is up to 2 orders of magnitude slower.
We may attribute this behaviour to ISEC's evaluation function which must compute a 1-tree for every node generated.
This effectively limits ISEC to solving much smaller instances than SEC.
Our goal however is not to be able to compute solutions faster than SEC but rather to illustrate the power of informed
heuristics for guiding ejection chain algorithms.
To that end we believe ISEC's ability to find alternative neighbourhoods which contain
demonstrably higher quality tours makes it a promising alternative.
Even so, we can report that the running time of all three algorithms can be improved
by artificially limiting the maximum length of each ejection chain from $2n$ to $n$. 
This change has a minimal effect on solution quality (the same trends were observed) but halves the CPU times 
in Table \ref{table-quality}.

\section{Conclusion}
We have shown that it is possible to generate Stem-and-Cycle ejection chains using an informed successor selection
strategy.
We develop the constrained 1-tree heuristic which computes a lower-bound on an apriori unknown goal tour with respect 
to a set of edge inclusion and edge exclusion constraints.
We show that this approach is both admissible and monotonic, never overestimating the cost to reach the goal tour.
We combine the constrained 1-tree heuristic with the Subpath Ejection Chain (SEC) procedure to develop a new informed
move operator which we term ISEC.
We show that ISEC, using only a single ejection chain, consistently computes higher quality solutions to the TSP 
than SEC which generates many more ejection chains. 
We give complexity results for ISEC and show that although its worst-case theoretical running time is $O(4n^4 \times
\log_{2}(n) + 4n^3)$ we are able to use it as a feasible alternative to SEC for TSP instances of up to 100 nodes.

Future work involves applying ISEC to similar size TSPs featuring precedence and time window constraints.
Often appearing as subproblems in other areas of combinatorial optimisation (for example the Vehicle Routing Problem
with Time Windows), these domains are interesting because the set of feasible solutions that may be found is much 
smaller than in the canonical TSP.
\subsection{Acknowledgements}
We would like to thank Adi Botea for his insightful comments and assistance during the
development of this work.
NICTA is funded by the Australian Government as represented by the Department of 
Broadband, Communications and the Digital Economy and the Australian Research 
Council through the ICT Centre of Excellence program.

\bibliographystyle{plain}
\bibliography{references}

\end{document}